\documentclass[10pt,twocolumn,letterpaper]{article}

\usepackage{cvpr}
\usepackage{times}
\usepackage{epsfig}
\usepackage{graphicx}
\usepackage{amsmath}
\usepackage{amssymb}

\usepackage[pagebackref=true,breaklinks=true,letterpaper=true,colorlinks,bookmarks=false]{hyperref}
\usepackage{graphicx}
\usepackage{amsthm}
\usepackage{algorithm}  
\usepackage{algorithmic}
\usepackage{epsfig} 
\usepackage{epstopdf}
\usepackage{color}
\usepackage{multirow}
\usepackage{subfigure}
\usepackage{placeins}
\usepackage{bm}
\usepackage{graphics}
\usepackage{footnote}
\usepackage{url}
\usepackage{indentfirst}
\usepackage{longtable}
\usepackage{array}
\usepackage{mdwmath}
\usepackage{mdwtab}
\usepackage{rotating}
\usepackage{color}
\usepackage{morefloats}
\usepackage{slashbox}
\usepackage{cite}
\usepackage{mcite}
\usepackage{eqnarray}
\usepackage{float}
\usepackage{makecell}
\usepackage{amsmath}
\usepackage{threeparttable}
\usepackage{pifont}
\usepackage{cleveref}
\usepackage{balance}

\renewcommand{\eqref}[1]{(\ref{#1})}

\crefname{equation}{}{}
\crefname{figure}{Fig.}{Figs.}
\crefname{table}{Table}{Tables}
\crefname{section}{Section}{Sections}
\crefname{prop}{Proposition}{Propositions}
\crefname{theorem}{Theorem}{Theorems}
\crefname{lemma}{Lemma}{Lemmas}
\crefname{algorithmic}{Algorithm}{Algorithms}
\crefname{algorithm}{Algorithm}{Algorithms}
\crefname{assumption}{Assumption}{Assumptions}

\newtheorem{theorem}{Theorem}[section]

\theoremstyle{plain}

\theoremstyle{definition}

 \cvprfinalcopy 

\def\cvprPaperID{3890} 

\ifcvprfinal\fi
\begin{document}

\title{ RES-PCA: A Scalable Approach to Recovering Low-rank Matrices }


\author{Chong Peng$^1$, Chenglizhao Chen$^1$, Zhao Kang$^2$, Jianbo Li$^1$, and Qiang Cheng$^3$ \\
$^1$College of Computer Science and Technology, Qingdao University \\
$^2$ School of Computer Science and Engineering, University of Electronic Science and Technology of China \\
$^3$ Department of Computer Science, University of Kentucky \\
{\tt\small \{pchong1991, cclz123\}@163.com, zkang@uestc.edu.cn, lijianbo@188.com, qiang.cheng@uky.edu}
}

\maketitle

\begin{abstract}
Robust principal component analysis (RPCA) has drawn significant attentions due to its powerful capability in recovering low-rank matrices as well as successful appplications in various real world problems. 
The current state-of-the-art algorithms usually need to solve singular value decomposition of large matrices, which generally has at least a quadratic or even cubic complexity.
This drawback has limited the application of RPCA in solving real world problems.
To combat this drawback, in this paper we propose a new type of RPCA method, 
RES-PCA, which is linearly efficient and scalable in both data size and dimension. 
For comparison purpose, AltProj, an existing scalable approach to RPCA requires the precise knowlwdge of the true rank; 
otherwise, it may fail to recover low-rank matrices.
By contrast, our method works with or without knowing the true rank;
even when both methods work, our method is faster. 
Extensive experiments have been performed and testified to the effectiveness of proposed method quantitatively and in visual quality, 
which suggests that our method is suitable to be employed as a light-weight, scalable component for RPCA in any application pipelines.

\end{abstract}


\section{Introduction}
\label{sec_intro}
Principal component analysis (PCA) bas been one of the most widely used techniques for unsupervised learning in various applications.
The classic PCA aims at seeking a low-rank approximation of a given data matrix.
Mathematically, it uses the $\ell_2$ norm to fit the reconstruction error, which is known to be sensitive to noise and outliers.
The harder problem of seeking a PCA effective for outlier-corruped data is called robust PCA (RPCA). 
There has been no mathematically precise meaning for the term ``outlier'' \cite{vaswani2018robust}.
Thus multiple methods have been attempted to define or quantify this term, such as alternating minimization \cite{ke2005robust}, 
random sampling techniques \cite{maronna2006robust,de2003framework}, 
multivariate trimming \cite{gnanadesikan1972robust}, 
and so on \cite{xu1995robust,croux2000principal}.

Among these methods, a recently emerged one treats an outlier as an additive sparse corruption \cite{wright2009robust},
which leads to decomposing the data into a low-rank and a sparse part.
Given data matrix $X\in\mathcal{R}^{d\times n}$, based on such a decomposition assumption, 
the corresponding RPCA method aims to mathematically solve the following problem \cite{wright2009robust,candes2011robust}:
\begin{equation}
\label{eq_rpca}
\min_{L,S} \text{rank}(L) + \lambda \|S\|_0,	\quad s.t.\quad X = L+S,
\end{equation}
where $\lambda\ge0$ is a balancing parameter, and $\|\cdot\|_{0}$ is the $\ell_0$ (pseudo) norm that counts the number of nonzero elements of the matrix. 
It is generally NP-hard to solve the rank function and $\ell_0$ norm-based optimization problems.
Hence, in practice \cref{eq_rpca} is often relaxed to the following convex problem \cite{candes2011robust}:
\begin{equation}
\label{eq_rpca_l1}
\min_{L,S} \|L\|_* + \lambda \|S\|_1,	\quad s.t.\quad X = L+S,
\end{equation}
where $\|\cdot\|_{*}$ is the nuclear norm that adds all singular values of the input matrix and $\|\cdot\|_{1}=\sum_{ij}|S_{ij}|$ is the $\ell_1$ norm of a matrix. 
A number of algorithms have been developed to solve \cref{eq_rpca_l1}, such as singular value thresholding (SVT) \cite{cai2010singular}, 
accelerated proximal gradient (APG) \cite{toh2010accelerated},
and inexact agumented Lagrange multipliers (IALM) \cite{lin2010augmented}. 
These algorithms, however, need to compute SVDs of matrices of size $d\times n$ at each iteration,
which, is known to generally have at least a quadratic or even cubic complexity \cite{golub2012matrix}. 
Thus, due to the use of SVDs, high complexity of these algorithms renders them less applicable to large-scale data.
To improve efficiency, an augmented Lagrange multipliers (ALM)-based algorithm adopts the PROPACK package \cite{ding2011bayesian} to solve partial, instead of full, SVDs. 
Even with partial SVD, it is still computationally costy when $d$ and $n$ are both large.

The convex RPCA in \cref{eq_rpca_l1} has two known limitations: 
1) Without the incoherence guarantee of the underlying matrix, or when the data is grossly corrupted, the results can be much deviated from the truth \cite{candes2011robust}; 
2) When the matrix has large singular values, its nuclear norm may lead to an estimation far from the rank \cite{kang2015robustpca}. 
To combact these drawbacks, several approaches to a better rank approximation have been proposed. 
For example, the rank of $L$ is fixed and used as a hard constraint in \cite{leow2013background}, 
and a nonconvex rank approximation is adopted to more accurately approximate the rank function in \cite{kang2015robustpca}.
However, these nonconvex approaches also need to solve full SVDs of $d\times n$ matrices.
Two methods in \cite{leow2013background,netrapalli2014non} need only to solve partial SVDs, 
which significantly reduces the complexity compared to full SVDs; 
for example, AltProj has a complexity of $O(r^2dn)$ \cite{netrapalli2014non}, with $r$ being the ground truth rank of $L$. 
However, if $r$ is not known a priori, \cite{netrapalli2014non} usually fails to recover $L$.

As large-scale data is increasingly ubiquitous, it is crucial to handle them with more efficient and scalable RPCA methods which, nonetheless, are still largely missing. 
To address such a need and challenge, in this paper, we propose a new RPCA method, called RES-PCA. 
This model does not depend on rank approximation to recover the low-rank component; 
rather, it effectivelly exploits the underlying group structural information of the low-rank component for the recovery. 
Consequently, the new method does not need to solve any SVDs as current state-of-the-art methods typically do, which avoids any quadratic or higher complexity; 
more specifically, the proposed method has a linear complexity in both $n$ and $p$, rendering it lightweight, scalable, and thus suitable for large-scale data applications.
We summarize the contributions of this paper as follows:
\begin{itemize}
\item We propose a new type of RPCA model exploiting the underlying group structures of the low-rank component.
\item We develop an ALM-based algorithm for optimization, which uses no matrix decomposition and has linearly efficient computation at each iteration. 
The new method is scalable in data dimension and sample size, suitable for large-scale data.
\item Extensive experiments have demonstrated the effectiveness of the proposed method quantitatively and qualitatively.
\end{itemize}


The rest of this paper is organized as follows. 
We first briefly review some related work. 
Then we introduce the new method and its optimization. 
Next, we conduct experiments to evaluate the new method. 
Finally, we conclude the paper.

\section{Related Work}
\label{sec_related}
The convex RPCA in \cref{eq_rpca_l1} considers the sparsity of the sparse component in an element-wise manner \cite{cai2010singular}. 
To exploit example-wise sparsity, the $\ell_{2,1}$ norm has been adopted by replacing the $\ell_{1}$ norm in \eqref{eq_rpca_l1} \cite{xu2010robust,mccoy2011two}:
\begin{equation}
\label{eq_rpca_l21}
\min_{L,S} \|L\|_* + \lambda \|S\|_{2,1},	\quad s.t.\quad X = L+S,
\end{equation}
where $\|S\|_{2,1} = \sum_{j}\sqrt{\sum_i S_{ij}^2}$ is the sum of $\ell_2$ norms of the columns. 
The difference between \cref{eq_rpca_l1,eq_rpca_l21} is that the latter incorporates spatial connections of the sparse component.

It is ponited out that the nuclear norm may be far from accurate in approximating the rank function \cite{peng2015subspace}. 
To alleviate this defficiency, some new rank approximations have been used to replace the nuclear norm in \cref{eq_rpca_l1,eq_rpca_l21},
such as $\gamma$-norm \cite{kang2015robustpca}.
The $\gamma$-norm based RPCA solves the following optimization problem:
\begin{equation}
\label{eq_rpca_capped}
\min_{L,S} \|L\|_{\gamma} + \lambda \|S\|_{2,1},	\quad s.t.\quad X = L+S,
\end{equation}
where $\|L\|_{\gamma}=\sum_i \frac{(1+\gamma)\sigma_i(L)}{\gamma+\sigma_i(L)}$, $\gamma > 0$, and $\sigma_i(L)$ is the $i$-th largest singular value of $L$. Here, with different values used for $\lambda$, the $\gamma$-norm may have different performance in approximating the rank function.  

Another recent nonconvex approach to RPCA, AltProj, cobmines the simplicity of PCA and elegant theory of convex RPCA \cite{netrapalli2014non}.
It alternatively projects the fitting residuals onto the low-rank and sparse sets. 
Given that the desired rank of $L$ is $r$, AltProj computes a rank-$k$ projection in each of the total $r$ stages,
with $k\in\{1,2,\cdots,r\}$. 
During this process, matrix elements with large fitting errors are discarded such that sparse errors are suppressed.
This method enjoys several nice properties; however, it needs the precise knowledge of the ground truth rank of $L$, which is not always available.
Without such knowledge, AltProj may fail to recover the low-rank component.

\section{New Robust PCA Method}
\label{sec_proposed}

The classic RPCA and its variants usually require to solve SVDs, which has a high complexity. 
To overcome this drawback, in this paper we consider a new type of RPCA model that has a linear complexity. 
Motivated by the convex RPCA approach, we assume that the data can be decomposed as $X \!=\! L \!+\! S$. 
Here, $L$ is the low-rank component of $X$ and its columns are linearly dependent in linear algebra;
hence, it is true that many columns of $L$ share high similarities and thus are close geometrically in Euclidean space.
In the case of a single rank-1 subspace, the above assumption naturally leads to the minimization of the sum of squared mutual distances, or equivalently the variance (scaled by $n$), of the column vectors of $L$: 
\begin{equation}
\label{eq_obj1}
\min_{L,S}	\lambda \sum_{i=1}^{n}\sum_{j=1}^{n}\|L_i - L_j\|_2^2 + \|S\|_1, s.t.\quad X = L+S,
\end{equation}
where $\lambda\ge 0$ is a balancing parameter, $L_i$ is the $i$th column of $L$, and $\|\cdot\|_2$ is the $\ell_2$ norm of a vector. 
It is noted that, though not necessary, it is sufficient that the minimization of the first term in \cref{eq_obj1} leads to low-rank structure for $L$.
To see this, we reformulate it as $2n\lambda \sum_{i=1}^{n} \|L_i - \frac{1}{n} \sum_{j=1}^{n} L_j\|_2^2$,
which is the sum of squares of residuals (SSR) from each data point to the average of all data points. 
Thus, by minimizing it, all columns are close to their average and the average is the minimizer of SSR, which ideally lead to rank-1 solution to $L$. Under some mild conditions, we have the following theorem.

\begin{theorem}
Given a matrix $L = [ {\textbf{l}}_1, \cdots, {\textbf{l}}_n ]$, with ${\textbf{l}}_i \in {\mathcal{R}}^p$, and   
$\| {\textbf{l}}_i \|_2^2 = s_i$, $i=1, \cdots, n$,
we have that $\mathrm{{rank}}(L) = 1$ is sufficient and necessary for 
\begin{equation}
\label{objective_fcn}
L = \underset{Q \in {\mathcal{R}}^{p \times n}}{{\mathrm{argmin}}} 
\textbf{Tr}(Q (I - \frac{1}{n} {\textbf{1}} {\textbf{1}}^T ) Q^T),
\end{equation}
s.t. 
 \begin{equation}
 \label{inequ_contraints}
 \|{\textbf{q}}_i \|_2^2 \le s_i , i=1, \cdots, n, 
 \end{equation}
 where $ Q = [{\textbf{q}}_1, \cdots, {\textbf{q}}_n ]$, and ${\textbf{1}}$ is an all-$1$ vector of dimension $n$. 
\end{theorem}

It is noted that the double summation in the first term of \cref{eq_obj1} can be written as $\textit{Tr}(L(I_n - \frac{1}{n}\textbf{1}\textbf{1}^T)L^T)$,
by minimizing which we can obtain the desired low-rank structure.
It is natural to generalize the above idea. To this end, we consider the case of multiple rank-1 subspaces with the following model, which we refer to as Robust, linearly Efficient, Scalable PCA (RES-PCA):
\begin{equation}
\label{eq_obj}
\begin{aligned}
& \min_{L,S,\{ p_1, \cdots, p_c \} }	\lambda  \sum_{i=1}^c \textit{Tr}
 \Big( L  \textbf{d}(p_i) \Big(I_n - \frac{1}{\|p_i\|_2}\textbf{1}_n\textbf{1}_n^T  \Big) \textbf{d}(p_i)  L^T \Big)	\\
&  + \|S\|_1 \quad	s.t.\quad\! X = L + S,\quad p_i\in\{0,1\}^{n},\quad \sum_i p_i = \textbf{1}_n,
\end{aligned}
\end{equation}
%
%
%
where $I_n$ is an identity matrix of size $n\times n$, $\textbf{1}_n$ is an $n$-dimensional column vector containing 1's, 
$\textbf{d}(\cdot)$ is an operator that returns a diagonal matrix from an input vector, 
and $p_i$ is a binary vector with the positions of $1$s indicating which of the $n$ column vectors belong to the $i$-th subspace. 
It is evident that by automatically learning $p_i$'s we are able to obtain the structural information about the low-rank subspaces.
It is noted that different norms can be used for $S$, such as $\ell_1$ and $\ell_{21}$ norms;
in this paper, without loss of generality, we adopt the $\ell_{1}$ norm to capture the sparse structure of $S$.
In next section, we will develop an efficient algorithm to optimize \cref{eq_obj}.

\textbf{Remark} In the case that data have nonlinear relationships, i.e., $L_i$ and $L_j$ are close on manifold rather than in Euclidean space if they come from the same subspace, a direct extension of our method can be made, which is presented in \cref{sec_opt_p}. 
Since the linear model provides with us the key ideas and contributions of this paper, and the experiments have confirmed its effectiveness in several real world applications, we focus on the linear model in our paper. Due to space limit, we do not fully expand the nonlinear model and will consider it in further research and more applications.

%

\section{Optimization}
In this section, we present an efficient ALM-based algorithm to solve \cref{eq_obj}. First, we define the augmented Lagrange function of \cref{eq_obj}:
\begin{equation}
\begin{aligned}
&	\mathcal{L} = \lambda \sum_{i=1}^c\textit{Tr} \Big( L  \textbf{d}(p_i) \Big(I_n - \frac{1}{\|p_i\|_2}\textbf{1}_n\textbf{1}_n^T  \Big) \textbf{d}(p_i)  L^T \Big) \\
& 	\qquad  + \|S\|_1 + \frac{\rho}{2}\| X - L - S + \frac{1}{\rho}\Theta \|_F^2	\\
&	s.t.\quad p_i\in\{0,1\}^n,\quad \sum_i p_i = \textbf{1}_n.
\end{aligned}
\end{equation}

Then we adpot the alternating decent approach to optimization, where at each step we optimize a subproblem with respect to a variable while keeping the others fixed. The detailed optimization strategies for each variable are described in the following.

\subsection{$L$-minimization}
The $L$-subproblem is to solve the following problem:
\begin{equation}
\label{eq_sub_L_1}
\begin{aligned}
&	\min_{L} \lambda \sum_{i=1}^c\textit{Tr} \Big( L  \textbf{d}(p_i) \Big(I_n - \frac{1}{\|p_i\|_2}\textbf{1}_n\textbf{1}_n^T  \Big) \textbf{d}(p_i)  L^T \Big) \\
& 	\qquad\qquad  + \frac{\rho}{2}\| X - L - S + \Theta / \rho \|_F^2	\\
\end{aligned}
\end{equation}

Omitting the factor $\lambda$, it is seen that the first term above can be derived as
\begin{equation}
\begin{aligned}
&	 \sum_{i=1}^c\textit{Tr} \Big( L  \textbf{d}(p_i) \Big(I_n - \frac{1}{\|p_i\|_2}\textbf{1}_n\textbf{1}_n^T  \Big) \textbf{d}(p_i)  L^T \Big) \\
=&   \sum_{i=1}^c\textit{Tr} \!\Big( \mathcal{P}_i(L) \Big( I_{\|p_i\|_2} - \frac{1}{\|p_i\|_2} \textbf{1}_{\|p_i\|_2}\textbf{1}_{\|p_i\|_2}^T \Big) \mathcal{P}_i^T(L) \Big),
\end{aligned}
\end{equation}
where the operator $\mathcal{P}_{i}(L)$ returns the submatrix of $L$ that contains the columns of $L$ corresponding to nonzeros of $p_i$. 
Correspondingly, it is straightforward to see that the second term of \cref{eq_sub_L_1} can be decomposed in a similar way:
\begin{equation}
\begin{aligned}
&  \frac{\rho}{2}\| X - L - S + \Theta / \rho \|_F^2	\\
= & \frac{\rho}{2} \sum_{i=1}^{c} \| \mathcal{P}_i(X - S + \Theta / \rho ) - \mathcal{P}_i(L) \|_F^2.
\end{aligned}
\end{equation}
Hence, $L$ can be solved by individually solving the following subproblems for $i=1,\cdots,c$:
%
%
\begin{equation}
\begin{aligned}
&\min_{\mathcal{P}_i(L)}\lambda \textit{Tr} \Big( \mathcal{P}_i(L) \Big( I_{\|p_i\|_2} - \frac{1}{\|p_i\|_2} \textbf{1}_{\|p_i\|_2}\textbf{1}_{\|p_i\|_2}^T \Big) \mathcal{P}_i^T(L) \Big)	\\
& \qquad + \frac{\rho}{2}\| \mathcal{P}_i(X - S + \Theta / \rho ) - \mathcal{P}_i(L) \|_F^2
\end{aligned}
\end{equation}
The above subproblems are convex and according to the first-order optimality condition we have 
\begin{equation}
\label{eq_sub_L_de_deriv}
\begin{aligned}
& 2\lambda \mathcal{P}_i(L) M_i + \rho \mathcal{P}_i(L) - \rho \mathcal{P}_i( D ) =0,
\end{aligned}
\end{equation}
where, for ease of presentation, we denote $D \!=\! X \!-\! S + \Theta / \rho$, and $M_i = I_{\|p_i\|_2} - \frac{1}{\|p_i\|_2} \textbf{1}_{\|p_i\|_2}\textbf{1}_{\|p_i\|_2}^T$. Hence, \cref{eq_sub_L_de_deriv} leads to the soluation of $\mathcal{P}_i(L)$:
\begin{equation}
\label{eq_sub_L_de_sol}
\begin{aligned}
&	\mathcal{P}_i(L) =  \rho (2\lambda  M_i + \rho I_{\|p_i\|_2})^{-1} \mathcal{P}_i( D ).
\end{aligned}
\end{equation}
It is seen that \cref{eq_sub_L_de_sol} requires matrix inversion, which, unfortunately, has a time complexity of $O(n^3)$ in general. To avoid matrix inversion, we re-write this matrix to simplify \cref{eq_sub_L_de_sol}:
\begin{equation}
\label{eq_sub_L_de_inv}
\!2\lambda  M_i + \rho I_{\|p_i\|_2} = (2\lambda + \rho) I_{\|p_i\|_2}  \!-\! \frac{2\lambda}{\|p_i\|_2} \textbf{1}_{\|p_i\|_2}\textbf{1}_{\|p_i\|_2}^T.
\end{equation}
It is notable that due to the special structure of \cref{eq_sub_L_de_inv} its inversion has a simple analytic expression by using the Sherman-Morrison-Woodbury formula:
%
\begin{equation}
\begin{aligned}
	&	\left( (2\lambda + \rho) I_{\|p_i\|_2} + (-\frac{2\lambda}{n}\textbf{1}_{\|p_i\|_2}) \textbf{1}_{\|p_i\|_2}^T \right)^{-1} \\
=	& \frac{1}{2\lambda + \rho} I_{\|p_i\|_2} \\
& \quad - \frac{\frac{1}{2\lambda + \rho} I_{\|p_i\|_2} (-\frac{2\lambda}{n}\textbf{1}_{\|p_i\|_2}) \textbf{1}_{\|p_i\|_2}^T \frac{1}{2\lambda + \rho} I_{\|p_i\|_2}}{1 + \textbf{1}_{\|p_i\|_2}^T \frac{1}{2\lambda + \rho} I_{\|p_i\|_2} (-\frac{2\lambda}{n}\textbf{1}_{\|p_i\|_2})}\\
=	&		\frac{1}{2\lambda + \rho} I_{\|p_i\|_2} + \frac{2\lambda}{ {\|p_i\|_2} \rho (2\lambda + \rho)} \textbf{1}_{\|p_i\|_2} \textbf{1}_{\|p_i\|_2}^T
\end{aligned}
\end{equation}
Hence, it is apparent that that \cref{eq_sub_L_de_sol} can be written as follows:
%
\begin{equation}
\begin{aligned}
&	\mathcal{P}_i(L) =  \Bigg( \frac{\rho}{2\lambda + \rho} I_{\|p_i\|_2} \\
& \qquad\qquad + \frac{2\lambda}{ {\|p_i\|_2} (2\lambda + \rho)} \textbf{1}_{\|p_i\|_2} \textbf{1}_{\|p_i\|_2}^T \Bigg) \mathcal{P}_i( D )	\\
& =	\frac{\rho}{2\lambda + \rho} \mathcal{P}_i( D ) \\
& \qquad\qquad + \frac{2\lambda}{ {\|p_i\|_2} (2\lambda + \rho)} (\textbf{1}_{\|p_i\|_2} (\textbf{1}_{\|p_i\|_2}^T \mathcal{P}_i( D ))),
\end{aligned}
\end{equation}
which has a linear complexity in both $n$ and $d$ by exploiting matrix-vector multiplications. $L$ can be obtained accordingly after obtaining all $\mathcal{P}_i(L)$, for $i=1,2,\cdots,c$. 


\subsection{$p_i$-minimization}
\label{sec_opt_p}
The subproblem associated with $p_i$-minimization is given as follows:
\begin{equation}
\begin{aligned}
&	\min_{p_i}	\sum_{i=1}^c\textit{Tr} \Big( L  \textbf{d}(p_i) \Big(I_n - \frac{1}{\|p_i\|_2}\textbf{1}_n\textbf{1}_n^T  \Big) \textbf{d}(p_i)  L^T \Big) \\
&	s.t.\quad  p_i\in\{0,1\}^{n},\quad \sum_i p_i = \textbf{1}_n.
\end{aligned}
\end{equation}
It is seen that 
\begin{equation}
\begin{aligned}
&	\sum_{i=1}^c\textit{Tr} \Big( L  \textbf{d}(p_i) \Big(I_n - \frac{1}{\|p_i\|_2}\textbf{1}_n\textbf{1}_n^T  \Big) \textbf{d}(p_i)  L^T \Big) \\
=&	\sum_{i=1}^c\textit{Tr} \Big( L   \Big(\textbf{d}(p_i) - \frac{1}{\|p_i\|_2} \textbf{d}(p_i)\textbf{1}_n\textbf{1}_n^T \textbf{d}(p_i) \Big)  L^T \Big) 
\nonumber\end{aligned}\end{equation}\begin{equation}\begin{aligned}
=&	\sum_{i=1}^c \| L   \Big(\textbf{d}(p_i) - \frac{1}{\|p_i\|_2} \textbf{d}(p_i)\textbf{1}_n\textbf{1}_n^T \textbf{d}(p_i) \Big) \| _F^2 \\
=& 	\sum_{i=1}^c \sum_{ (p_i)_j = 1 } \| L_j - \frac{1}{ \|p_i\|_2}\sum_{ (p_i)_j = 1 } L_j \|_2^2,
\end{aligned}
\end{equation}
where $(p_i)_j$ denotes the $j$-th element of $p_i$. 
Hence, the $p_i$-subproblems can be converted to 
\begin{equation}
\label{eq_sub_p}
\begin{aligned}
&	\min_{p_i}	\sum_{i=1}^c \sum_{ (p_i)_j = 1 }\| L_j - \frac{1}{ \|p_i\|_2}\sum_{ (p_i)_j = 1 } L_j \|_2^2  \\
&	s.t.\quad  p_i\in\{0,1\}^{n},\quad \sum_i p_i = \textbf{1}_n,
\end{aligned}
\end{equation}
which is simply the standard K-means problem. This is surprising in that we only need to perform K-means to $L$ and then the optimal $[p_1,\cdots,p_c]\in\{0,1\}^{n\times c}$ simply corresponds to the group indicator matrix:
\begin{equation}
[p_1,\cdots,p_c] \leftarrow \textit{K-means} (L,c).
\end{equation}
It should be noted that with its current form, \cref{eq_sub_p} is solved by K-means \cite{Peng:2018:ICD:3210369.3200488}. However, more general clustering methods can be also applicable if we consider solving $p_i$ as a clustering rather than optimization problem.
For example, if we consider nonlinear clustering algorithms, such as spectral clustering, the recovered $L$ and $p$ actually reflect nonlinear structures of the data,
which can be treated as a direct nonlinear extension of our method to account for nonlinear relationships of the data. 

\subsection{$S$-minimization}
The $S$-subproblem is
\begin{equation}
\begin{aligned}
\min_{S} \frac{1}{\rho}\|S\|_1 + \frac{1}{2}\| X - L - S + \Theta / \rho \|_F^2,
\end{aligned}
\end{equation}
which is solved using the soft-thresholding operator \cite{beck2009fast,daubechies2004iterative}:
\begin{equation}
S_{ij}=\left(\left| B_{ij}\right| - 1 / \rho \right)_{+}\textit{sign}\left(B_{ij}\right),
\end{equation}
where $B = X - L + \Theta / \rho.$

\subsection{$\Theta,\rho$-updating}
For the updating of $\Theta$ and $\rho$, we follow a standard approach in ALM framework:
\begin{equation}
\begin{aligned}
\Theta &	= \Theta + \rho(X - L - S),	\\
\rho &	= \rho \kappa,
\end{aligned}
\end{equation}
where $\kappa >1$ is a parameter that controls the increasing speed of $\rho$.

Regarding the complexity of the above optimization procedure, it should be noted that each step requires $O(nd)$ complexity and typically ALM converges in a finite number of steps \cite{boyd2011distributed}, thus the overall complexity of our method is $O(nd)$.

\section{Experiments}
In this section, we evaluate the proposed method in comparison with several current state-of-the-art algorithms, including variational Bayesian RPCA (VBRPCA) \cite{ding2011bayesian}, IALM for convex RPCA \cite{candes2011robust}, AltProj \cite{netrapalli2014non}, NSA \cite{aybat2011fast}, and PCP \cite{zhou2010stable}.
In particular, we follow \cite{peng2016fast,kang2015robustpca} and evaluate RES-PCA in three applications, including foreground-background separation from video sequences, shadow removal from face images, and anamoly detection from hand-written digits.
All these experiments are conducted under Ubuntu system with 12 Intel(R) Xeon(R) W-2133 CPR \@3.60GHz.
All algorithms are terminated if a maximum of 500 iterations is reached or 
$\max\{\frac{\|X-L_{t}-S_{t}\|_F}{\|X\|_F}, \frac{\|L_{t+1} - L_{t}\|_F}{\|X\|_F}, \frac{\|\|S_{t+1} - S_{t}\|_F}{\|X\|_F}\}\le 0.001$ is satisfied.

\subsection{Foreground-Background Seperation}

Foreground-background separation is to detect moving objects or interesting activities in a scene, and remove background(s) from a video sequence. 
The background(s) and moving objects correspond to the low-rank and sparse parts, respectively. 
For this task, we use 9 datasets, whose characteristics are summarized in \cref{tab_datadescription}. 
Among these video datasets, the first 5 contain a single background while the remaining sequences have 2 backgrounds.


\begin{table}[!tb]
\Large
\centering
\caption{ Description of Video Sequence Data Sets }
\resizebox{0.9\columnwidth}{!}{
\begin{tabular}{ |c |c |c | }
\hline		
Data Set			& data size							& \# of backgrounds		\\ 	\hline	
Escalator Airport	& 130$\times$160 $\times$ 3,417		& 1		\\	\hline
Hall Airport		& 144$\times$176 $\times$ 3,584		& 1		\\	\hline
Bootstrap 			& 120$\times$160 $\times$ 2,055		& 1		\\	\hline	
Shopping Mall 		& 256$\times$320 $\times$ 1,286		& 1		\\	\hline
Highway		 		& 240$\times$320 $\times$ 1,700		& 1		\\	\hline\hline

Lobby	 			& 128$\times$160 $\times$ 1,546		& 2		\\	\hline
Camera Parameter	& 240$\times$320 $\times$ 5,001		& 2		\\	\hline
Light Switch-1		& 120$\times$160 $\times$ 2,800		& 2		\\	\hline
Light Switch-2		& 120$\times$160 $\times$ 2,715		& 2		\\	\hline
\end{tabular}
} 
\begin{flushleft}
\end{flushleft}
\label{tab_datadescription}
\end{table}

\begin{table}[!tb]
\Large
\centering
\caption{ Results of Foreground-Background Separation }
\resizebox{1\columnwidth}{!}{
\begin{tabular}{|c||c|| c |c |c |c| c| }
\hline		
Data 	
& Method	& Rank($L$) & ${\|S\|_0}/{(d n)}$ & $\frac{\|X-L-S\|_F}{\|X\|_F}$	& \# of Iter. 	& Time	\\ \hline	


\multirow{5}{1.8cm}{ Bootstrap } 	
& AltProj	& 1			& 0.9397		& 4.22e-4		& 36			& 68.61		\\	\cline{2-7}
& NSA		& 843		& 0.7944		& 5.87e-4		& 12			& 1343.22	\\	\cline{2-7}
& VBRPCA	& 1			& 1.0000		& 9.90e-4		& 175			& 186.90	\\	\cline{2-7}
& IALM		& 782		& 0.8003		& 6.11e-4		& 15			& 1356.04	\\	\cline{2-7}
& PCP		& 1174		& 0.7859		& 3.45e-4		& 94			& 571.75	\\	\cline{2-7}
& RES-PCA	& 1			& 0.9379		& 7.81e-4		& 23			& 16.73		\\	\hline\hline


\multirow{5}{1.8cm}{ Escalator Airport } 	
& AltProj	& 1			& 0.8987		& 3.86e-4		& 33			& 69.34		\\	\cline{2-7}
& NSA		& 1016		& 0.6390		& 8.09e-4		& 12			& 1793.35	\\	\cline{2-7}
& VBRPCA	& 1			& 0.9839		& 9.76e-4		& 134			& 168.01	\\	\cline{2-7}
& IALM		& 1065		& 0.6482		& 6.95e-4		& 15			& 1325.40	\\	\cline{2-7}
& PCP		& 1232		& 0.6670		& 3.59e-4		& 93			& 727.65	\\	\cline{2-7}
& RES-PCA	& 1			& 0.8898		& 5.77e-4		& 23			& 20.47		\\ 	\hline	\hline



\multirow{5}{1.8cm}{ Hall Airport } 	
& AltProj	& 1			& 0.9573		& 1.69e-5		& 37			& 93.62		\\	\cline{2-7}
& NSA		& 948		& 0.7489		& 4.89e-4		& 13			& 2189.99	\\	\cline{2-7}
& VBRPCA	& 1			& 1.0000		& 9.90e-4		& 152			& 240.17	\\	\cline{2-7}
& IALM		& 974		& 0.6917		& 7.37e-4		& 14			& 2024.10	\\	\cline{2-7}
& PCP		& 1292		& 0.7055		& 4.27e-4		& 77			& 744.28	\\	\cline{2-7}
& RES-PCA	& 1			& 0.9302 		& 5.82e-4		& 23			& 26.38		\\	\hline\hline

\multirow{5}{1.8cm}{ Highway } 	
& AltProj	& 1			& 0.8846		& 4.63e-4		& 27			& 119.17	\\	\cline{2-7}
& NSA		& 166		& 0.9732		& 0.87e-4		& 15			& 1238.95	\\	\cline{2-7}
& VBRPCA	& 1			& 1.0000		& 9.87e-4		& 126			& 287.27	\\	\cline{2-7}
& IALM		& 357		& 0.7980		& 6.25e-4		& 15			& 1409.10	\\	\cline{2-7}
& PCP		& 531		& 0.8440		& 2.27e-4		& 152			& 1013.00	\\	\cline{2-7}
& RES-PCA	& 1			& 0.9340 		& 7.20e-4		& 23			& 35.32		\\	\hline\hline


\multirow{5}{1.8cm}{ Shopping Mall } 	
& AltProj	& 1			& 0.8907		& 8.12e-4		& 30			& 85.92		\\	\cline{2-7}
& NSA		& 174		& 0.9372		& 1.57e-4		& 14			& 1027.45	\\	\cline{2-7}
& VBRPCA	& 1			& 1.0000		& 9.92e-4		& 157			& 295.00	\\	\cline{2-7}
& IALM		& 151		& 0.8457		& 6.25e-4		& 14			& 498.65	\\	\cline{2-7}
& PCP		& 290		& 0.8898		& 2.85e-4		& 165			& 790.30	\\	\cline{2-7}
& RES-PCA	& 1			& 0.9208		& 7.94e-4		& 23			& 28.44		\\	\hline\hline


\multirow{5}{1.8cm}{Lobby} 	
& AltProj	& 2			& 0.88.97		& 3.77e-4		& 26			& 21.58		\\	\cline{2-7}
& NSA		& 161		& 0.8073		& 6.13e-4		& 13			& 182.50	\\	\cline{2-7}
& VBRPCA	& 2			& 1.0000		& 9.92e-4		& 111			& 69.47		\\	\cline{2-7}
& IALM		& 104		& 0.8229		& 5.66e-4		& 15			& 168.22	\\	\cline{2-7}
& PCP		& 502		& 0.8500		& 2.59e-4		& 92			& 166.79	\\	\cline{2-7}
& RES-PCA	& 2			& 0.8963		& 1.83e-4		& 25			& 20.11		\\	\hline\hline

\multirow{5}{1.8cm}{Camera Parameter} 	
& AltProj	& ------	& ------		& ------		& ------		& ------	\\	\cline{2-7}
& NSA		& ------	& ------		& ------		& ------		& ------	\\	\cline{2-7}
& VBRPCA	& 1			& 1.0000		& 9.95e-4		& 171			& 1108.20	\\	\cline{2-7}
& IALM		& 1123		& 0.7020		& 7.81e-4		& 16			& 9297.40	\\	\cline{2-7}
& PCP		& ------	& ------		& ------		& ------		& ------ 	\\	\cline{2-7}
& RES-PCA	& 2			& 0.8305		& 2.48e-4		& 25			& 303.57	\\	\hline\hline

\multirow{5}{1.8cm}{Light Switch-1} 	
& AltProj	& 2			& 0.90.84		& 4.21e-4		& 48			& 73.54		\\	\cline{2-7}
& NSA		& 541		& 0.6559		& 5.87e-4		& 13			& 687.19	\\	\cline{2-7}
& VBRPCA	& 1			& 1.0000		& 9.83e-4		& 165			& 151.05	\\	\cline{2-7}
& IALM		& 415		& 0.6298		& 9.21e-4		& 14			& 496.92	\\	\cline{2-7}
& PCP		& 848		& 0.6776		& 5.91e-4		& 85			& 410.39	\\	\cline{2-7}
& RES-PCA	& 2			& 0.9708		& 4.15e-4		& 23			& 31.68		\\	\hline\hline

\multirow{5}{1.8cm}{Light Switch-2} 	
& AltProj	& 2			& 0.8078		& 9.01e-4		& 37			& 44.34		\\	\cline{2-7}
& NSA		& 486		& 0.8041		& 4.90e-4		& 14			& 846.81	\\	\cline{2-7}
& VBRPCA	& 1			& 1.0000		& 9.93e-4		& 150			& 141.21	\\	\cline{2-7}
& IALM		& 333		& 0.7815		& 7.79e-4		& 15			& 616.28	\\	\cline{2-7}
& PCP		& 985		& 0.8337		& 2.68e-4		& 154			& 756.34	\\	\cline{2-7}
& RES-PCA	& 2			& 0.8608		& 2.82e-4		& 25			& 33.71		\\	\hline

\end{tabular}
} 
{\\ \scriptsize We set the rank to be the minimal number of singular values that contribute more  \\
  than 99.5\% information to avoid the noise effect of small singular values. \\
 ``------'' presents an ``out of memory'' issue. \\}
\label{tab_num_exact}
\end{table}

For the parameters, we set them as follows. For IALM, we use the theoretically optimal balancing parameter $\frac{1}{\sqrt{\max{(n,d)}}}$. 
The same balancing parameter is used for PCP and NSA as suggested in the original papers. 
For fair comparison, we use $\sqrt{\max{(n,d)}}$ for the proposed method. 
For AltProj, we specify the ground truth rank; for VBRPCA, we use the ground truth rank as its initial rank parameter. 
For fair comparison, we set $c$ to be ground truth rank for RES-PCA. 
For all methods that relay on ALM-optimization, we set the parameters to be $\rho = 0.0001$ and $\kappa = 1.5$. 
These settings remain the same throughout this paper unless specified otherwise.

We show the results in \cref{tab_num_exact}. 
It is observed that AltProj, VBRPCA, and RES-PCA are able to recover the backgrounds from the video with low rank while IALM, NSA and PCP with much higher ranks. 
However, it is noted that VBRPCA may recover $L$ with ranks lower than the ground truth. 
For example, on Light Switch-1, Light Switch-2, and Camera Parameter data sets, the ground truth rank of the background is 2 whereas VBRPCA recovers the low rank parts with rank 1. 
This may be a potential problem, as will be clear later on in visual illustration.
Although IALM, NSA amd PCP do not recover $L$ with desired low ranks, they recovery $S$ more sparsely than AltProj, VBRPCA, and RES-PCA. 
Besides, we observe that the speed of the proposed method is superior to that of the other methods. 
From \cref{tab_num_exact}, it is observed that the proposed method is about 3 times faster than AltProj, 
the second fastest one, and more than 10 (even about 60 on some data sets) times faster than IALM. 
Although the proposed method does not obtain the smallest errors at convergence on some data, 
it is noted that the levels of the errors are well comparable to the other methods.

\begin{figure*}[!tb]
\centering
\resizebox{0.95\textwidth}{!}{
	{\includegraphics[width=\textwidth]{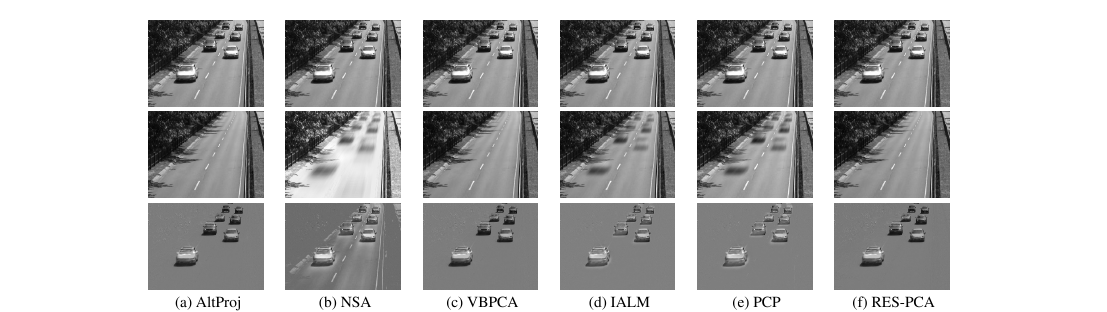} }
%
%
%
%
}
\caption{ Foreground-background separation in the Highway video. The top to the bottom are the original frame, extracted background, and foreground, respectively.  }
\label{fig_highway}
\end{figure*}

\begin{figure*}[!tb]
\centering
\resizebox{0.95\textwidth}{!}{
	{\includegraphics[width=\textwidth]{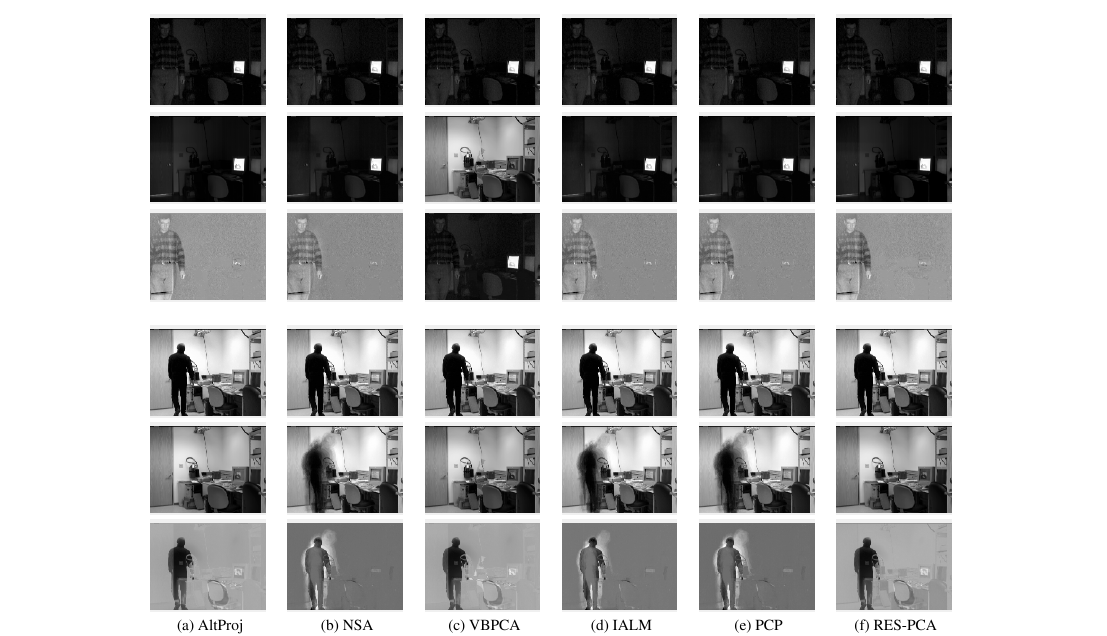} }
}
\caption{  Foreground-background separation in the Light Switch-2 video. Within the first and last 3 lines, the top to bottom are the original frame, extracted background, and foreground, respectively.   }
\label{fig_lightswitch}
\end{figure*}

It should be noted that for mthods such as IALM, PCP, and NSA, though they do not recover $L$ with desired low ranks, 
it is possible that by tunning their balancing parameters they may work well.
However, tunning parameter for unsupervised learning method is usually time consuming. 
The proposed method has one balancing parameter, which has been empirically verified that the theoretical parameter as provided in \cite{candes2011robust} works well. 
A possible explaination is that RES-PCA has a close connection and thus enjoies the same optimal parameter with the convex RPCA.
More theoretical validation is to be explored in further work.

Moreover, to visually compare the algorithms and illustrate the effectiveness of the proposed method, 
we show some decomposition results in \cref{fig_highway,fig_lightswitch}.
Since IALM, NSA and PCP cannot recover $L$ with desired low ranks, they cannot recover the backgrounds well. 
For example, we can observe shadows of car on highway in \cref{fig_highway}.
VBRPCA reocvers $L$ with ranks lower than the ground truth on some data sets; 
consequently, on such data as Light Switch-2 in \cref{fig_lightswitch} we can see that VBRPCA cannot work well on data with different backgrounds.
AltProj and RES-PCA can separate the backgrounds and foregrounds well.

\begin{figure}[h]
\centering
\resizebox{1\columnwidth}{!}{
	{\includegraphics[width=\columnwidth]{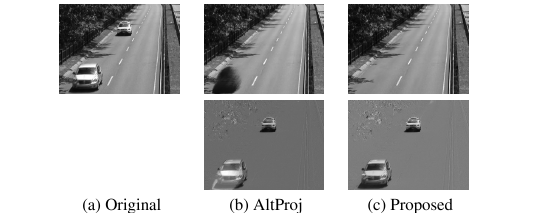} }
%
%
}
\caption{ Foreground-background separation in the Highway video when the ground truth rank is unknown and, consequently, $c$ is specified to a wrong value. The top left is the original frame and the rest are extracted background (top) and foreground (bottom).  }
\label{fig_highway5}
\end{figure}

\begin{figure}[h]
\centering
\large
\resizebox{1\columnwidth}{!}{
	{\includegraphics[width=\columnwidth]{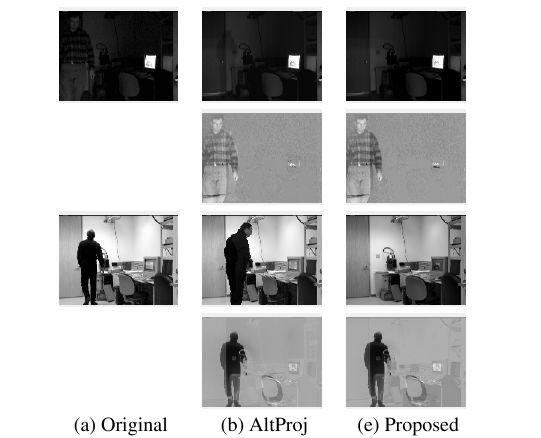} }
%
%
%
%
%
}
\caption{ Foreground-background separation in the Light Switch-2 video. Within the two and bottom two rows, the top left is the original frame and the rest are extracted background (top) and foreground (bottom), respectively.  }
\label{fig_lightswitch5}
\end{figure}

To further assess the performance of the proposed method, 
we conduct the following experiments to compare the two methods that have achieved the top performance: AltProj and RES-PCA. 
In this test we asume that the ground truth rank of $L$ is unknown, and we set it to 5 for AltProj and $c=5$ for the proposed method. 
Some obtained results are given in \cref{fig_highway5,fig_lightswitch5}. 
It is seen that RES-PCA can still separate the background and foreground well while AltProj fails.
The success of RES-PCA in this kind of scenarios can be explained as follows: 
With $c$ greater than the ground truth rank of $L$,
a large group of backgrounds is usually divided into smaller groups such that the backgrounds within each group still share the same structure;
as a consequence, RES-PCA can still recover the low-rank matrices correctly. 
This observation reveals that RES-PCA has superior performance to AltProj when the precise knowledge of the ground truth is unknown a priori.

%
%

\subsection{ Shadow removal from face images }
\begin{figure}[h]
\centering
\resizebox{1.0\columnwidth}{!}{
	{\includegraphics[width=\textwidth]{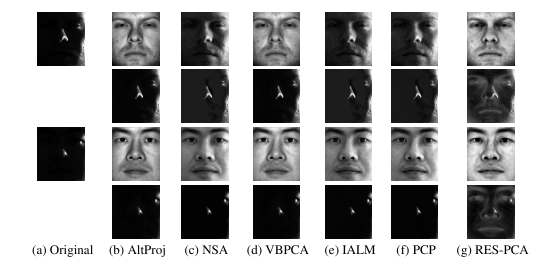} }
}
\caption{ Shadow removal from face images. Every two rows are results for different original faces. For each original face, the first row are shadow removed face images, while the second row are shadow images.  }
\label{fig_subject1}
\end{figure}

Face recognition is an important topic; 
however, it is often plagued by heavy noise and shadows on face images \cite{basri2003lambertian}. 
Therefore, there is a need to handle shadows.
In this test, low-rank methods are used because the (unknown) clean images reside in a low-rank subspace, 
corresponding to $L$, while the shadows correspond to $S$.
We use the Extended Yale B (EYaleB) data set for comparative study. 
EYaleB data contains face images from 38 persons, among which we select images of the first 2 persons, 
namely, subject 1 and subject 2. 
For each there are 64 images of $192\times 168$ pixels. 
Following the common approach as in \cite{candes2011robust,kang2015robustpca}, we construct a data matrix for each person by vectorizing the images and perform different RPCA algorithms on the matrix. 
We show some results in \cref{fig_subject1} for visual inspection. 
It is observed that all methods can successfully remove shadows on subject 2, but some fail on subject 1. 
The proposed method removes shadows from face images on both subject 1 and subject 2, which confirms its effectiveness.

\subsection{Anomaly Detection}
\begin{figure}[!tb]
\centering
\includegraphics[width=1.0\columnwidth]{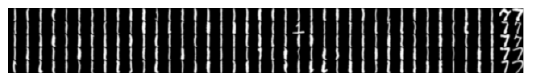} 
\caption{Selected `1's and `7's from USPS dataset. }
\label{fig_usps_example}
\end{figure}

\begin{figure}[h]
\centering
\includegraphics[width=1.0\columnwidth]{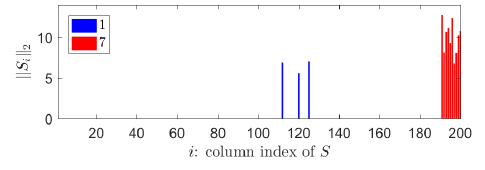}  
\caption{Values of $\|S_i\|_2$. }
\label{fig_usps_s}
\end{figure}

\begin{figure}[h]
\centering
\includegraphics[width=0.9\columnwidth]{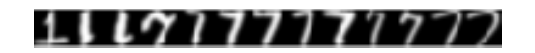} 
\caption{Detected outliers from the data set. }
\label{fig_detected}
\end{figure}

Given a number of images from a subject, they form a low-dimensional subspace. 
Those images with stark differences from the majority can be regarded as outliers; 
besides, a few images from another subject are also treated as outliers. 
Anomaly detection is to identify such kinds of outliers from the dominant images. 
It is modeled that $L$ is comprised of the dominant images while $S$ captures the outliers.
For this test, we use USPS data set which consists of 9,298 hand-written digits of size $16\times 16$. 
We follow \cite{kang2015robustpca} and vectorize the first 190 images of `1's and the last 10 of `7's to construct a $256\times 200$ data matrix. 
Since the dat set contains much more `1's than `7's, we regard the former as the dominant digit while the latter outlier.
For visual illustration, we show examples of these digit images in \cref{fig_usps_example}. 
It is observed that all the `7's are outliers. Besides, some `1's are quite different from the majority, such as the one with an underline. 
We apply RES-PCA to this data set and obtain the separated $L$ and $S$. 
In $S$, those columns corresponding to outliers have relatively larger values.
Following \cite{kang2015robustpca}, we use the $\ell_2$ norm to measure the columns of $S$ and show their values in \cref{fig_usps_s}, 
where we have vanished values smaller than 5 for clearer visualization.
Then we show the corresponding digits in \cref{fig_detected}, which are the detected outliers.
It is noted that RES-PCA has detected all the `7's as well as some `1's, such as the one with an underline. 
This has verified the effectiveness of RES-PCA in anomaly detection.

\subsection{Scalability}
\begin{figure}[h]
\centering
\includegraphics[width=0.98\columnwidth]{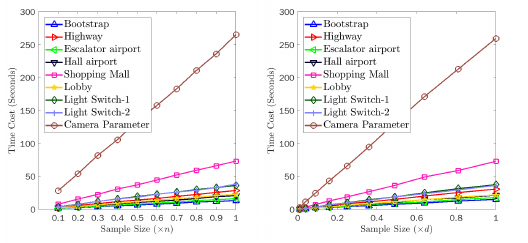} 
\caption{Time cost with respect to $n$ and $d$ on different data sets (best viewed in color). }
\label{fig_scalability}
\end{figure}

We have analyzed the scalability of the proposed method in previous sections. 
In this test, we empirically verify the result from our analysis regarding the linearity with $n$ and $d$ using the data sets in \cref{tab_datadescription}. 
For each of these data sets, we use different sampling ratios in sample size and data dimension, respectively, to collect its subsets of different sizes.
On each subset, we perform RES-PCA 10 times. From \cref{tab_num_exact}, it is seen that all experiments are terminated within about 23-25 iterations; hence, in this test we temporarily ignore the terminating tolerance and terminate the experiment within a reasonable number of iterations, which is set to be 30.
Then we report the average time cost and show the results in \cref{fig_scalability}.
It is observed that the time cost of RES-PCA increases linearly in both $n$ and $d$, which confirms the scalability of the proposed method. 

%

\section{Conclusion}
Existing RPCA methods typically need to solve SVDs of large matrices, which generally has at least a quadratic or even cubic complexity. 
To combat this drawback, in this pape we propose a new type of RPCA method.
The new method recovers the low-rank component by exploiting geometrical similarities of the data,
without performing any SVD that current state-of-the-art RPCA methods usually have to do.
We develop an ALM-based optimization algorithm which is linearly efficient and scalable in both data dimension and sample size.
Extensive experiments in different applications testify to the effectivenss of the proposed method, 
in which we observe superior performance in speed and visual quality to several current state-of-the-art methods.
These observations suggest that the proposed method is suitable for large-scale data applications in real world problems.

\section*{Acknowledgement}
This work is supported by National Natural Science Foundation of China under grants 61806106, 61802215, 61806045, 61502261, 61572457, and 61379132. C. Chen and Q. Cheng are corresponding authors.

{ \small
\bibliographystyle{ieee_fullname}
\bibliography{rpca_cvpr_final}
}

\end{document}